# Automatically Predict Material Properties with Microscopic Image: Example Polymer Miscibility

Zhilong Liang, Zhenzhi Tan, Ruixin Hong, Wanli Ouyang, Jinying Yuan[*] and Changshui Zhang[*]

Corresponding Authors

Jinying Yuan - Key Lab of Organic Optoelectronics and Molecular Engineering of Ministry of Education; Department of Chemistry, Tsinghua University, Beijing, P. R. China. Email: yuanjy@tsinghua.edu.cn

Changshui Zhang - Institute for Artificial Intelligence, Tsinghua University (THUAI); Beijing National Research Center for Information Science and Technology (BNRist); Department of Automation, Tsinghua University, Beijing, P. R. China. Email: zcs@mail.tsinghua.edu.cn

Authors

Zhilong Liang - Institute for Artificial Intelligence, Tsinghua University (THUAI); Beijing National Research Center for Information Science and Technology (BNRist); Department of Automation, Tsinghua University, Beijing, P. R. China.

Zhenzhi Tan - Department of Chemistry, Tsinghua University, Beijing, P. R. China.

Ruixin Hong - Institute for Artificial Intelligence, Tsinghua University (THUAI); Beijing National Research Center for Information Science and Technology (BNRist); Department of Automation, Tsinghua University, Beijing, P. R. China.

Wanli Ouyang - Shanghai AI Lab, Shanghai, P. R. China. Email: ouyangwanli@pjlab.org.cn

**Abstract**: Many material properties are manifested in the morphological appearance and characterized with microscopic image, such as scanning electron microscopy (SEM). Polymer miscibility is a key physical quantity of polymer material and commonly and intuitively judged by SEM images. However, human observation and judgement for the images is time-consuming, labor-intensive and hard to be quantified. Computer image recognition with machine learning method can make up the defects of artificial judging, giving accurate and quantitative judgement. We achieve automatic miscibility recognition utilizing convolution neural network and transfer learning method, and the model obtains up to 94% accuracy. We also put forward a quantitative criterion for polymer miscibility with this model. The proposed method can be widely applied to the quantitative characterization of the microstructure and properties of various materials.

## 1. Introduction

The methods to computing and characterizing material properties have developed rapidly in recent years. In terms of material property calculations, methods such as molecular dynamics (MD) simulations achieved good results in free energy calculation[1, 2] and drug discovery[3, 4]. In addition to material property calculation, material property characterization is also a crucial part of material researches. Among kinds of characterization methods, scanning electron microscopy (SEM) is one of the most important means to determining macroscopic and intrinsic properties of the materials. At the present stage, this examination technology relies heavily on human observation and judgement, which is time-consuming, labor-intensive and hard to be quantified.

Recently, the development of artificial intelligence and machine learning makes automatic image recognition applicable[5]. With the help of computer image recognition technology, we could build an effective bridge from the microscopic morphology to the macroscopic properties. Allen et al.[6] used convolution neutral network (CNNs) to identify the common functional groups in gas-phase FTIR (Fourier transform infrared spectroscopy), simplified the recognition process of infrared spectrum. Lee et al.[7] developed a GA-based image analysis method to accurately count the number of nanoparticles and achieved 99.75% accuracy. Gao et al.[8] obtained the function model of material $T_g$ and image characteristics through machine learning by extracting microscopic image features of materials. Other researchers also have tried to use machine learning methods to analyze SEM and other microscopic images of materials, including morphological category [9-11], particle size [12], identification and statistics of particle numbers [13]. We want to further explore the relationship between SEM images and material properties.

However, directly bridging microscopic images and material macroscopic properties is still challenging. Considering that polymer blends are important engineering materials and that their miscibility is a key index to determining their performance, we chose quantitative judgment of polymer miscibility with SEM images as an example. Polymer blend materials involve blending two or more polymers to produce a new material. These blends have been extensively researched due to their capacity to combine the advantageous properties of different polymers and their ease of synthesis and processing. Researchers have utilized machine learning methods to investigate their properties, including the mechanical properties[14], liquid crystal behavior[15], thermal conductivity[16], dielectric constant[17], optical properties[18] and molecular design[19]. Among all kinds of properties, polymer miscibility is the primary basis for determining the structure and properties of blends[20], and it is difficult for immiscible polymers to form materials with good properties. Therefore, determining whether different polymers are miscible is a crucial question and attracts researchers' focus[21].

While the miscibility of some blends needs to be determined through technologies in physical chemistry such as differential scanning calorimetry (DSC), microscopic observation (such as SEM) is the most efficient approach since this property can be manifested in the homogeneity of morphological appearance at most times[22, 23]. For example, a homogeneous phase is one of the features of miscible polymers, while immiscible polymers usually present clear phase separation and form a spherical dispersion phase or sea-island structure. Typical SEM images of polymer blends are shown in Fig.1. Figure A[24] is the SEM image of a miscible polymer blend, in which two polymers form a homogeneous phase. Figure B[25] is the SEM image of an immiscible polymer blend and presents clear phase separation. As previously mentioned, determination of polymer miscibility remains dependent on researcher judgement and lacks a set of accurate and quantitative technical means to describe polymer miscibility. In this work, we present an automated judging approach to improve efficiency and accuracy.

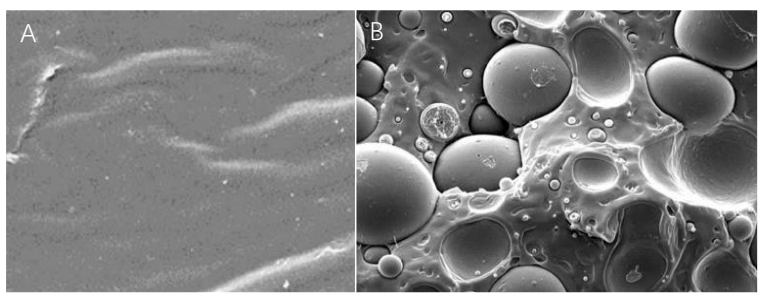

Figure 1. Typical SEM images of polymer blends.

A: miscible polymer blend. Adapted with permission from reference 24. Copyright 2009, Springer Science Business Media B.V.

B: immiscible polymer blend. Adapted with permission from reference 25. Copyright 2008 Elsevier Ltd.

We designed a deep neural network model to realize miscibility recognition with microscopic images of polymer blends. Also, we used transfer learning, a branch of machine learning to improve the prediction accuracy of the model. We collected 517 SEM images of polymers in total from the literature, combined with the PoLyInfo database[26] and miscibility descriptions in the literature. Our model is pretrained in ImageNet[27](A database for object classification) and then finetuned with the SEM images of polymer blends, which is so-called transfer learning. The model can give the judgment of polymer miscibility with 94% accuracy, making it possible to obtain the quantitative miscibility judgment index and establish a quantitative criterion of polymer miscibility. Our main contributions are:

(1) A machine learning framework (Figure 2) is proposed to bridge material macroscopic properties and microscopic images. In the miscibility recognition task, we achieved 94% accuracy, which is comparable to the determination of researchers. It automates the morphology analysis of polymer blending materials and facilitates the development of high-throughput experiments.

(2) A specific polymer miscibility image database is built, including SEM images, original descriptions in literature, and the microscopic scale. It will help researchers study polymer blend morphology and conduct downstream tasks.

(3) A quantitative criterion for polymer miscibility is established. This quantitative description will facilitate to study other properties of polymer blend related to miscibility.

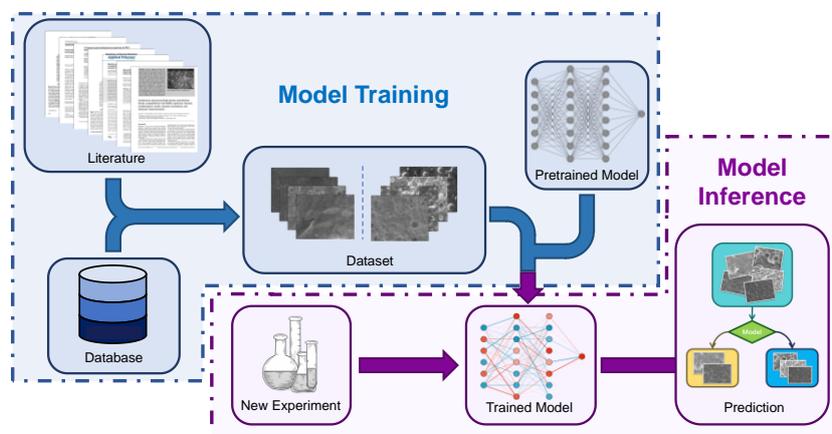

Figure 2. The machine learning framework to bridge material macroscopic properties and microscopic images

## 2. Methods

**2.1.** Dataset

Dataset plays a crucial important role in machine learning and needs to be accurate[28]. We extracted microscopic images mainly according to the PolyInfo[26] database and literature descriptions. PolyInfo is a polymer database from NIMS that provides data for the design of various polymer materials. The morphology, characteristics of material, mixing or blending method, component ratio, and references of polymer blends were collected from the PolyInfo database. Common microscopic images include scanning electron microscopy (SEM) and transmission electron microscopy (TEM) images. Through our collection, the SEM image dataset has a larger size than the TEM (transmission electron microscope) dataset. As the size of the database is a crucial factor in machine learning and SEM images are more accessible than TEM images, we focus our current work on SEM images. Totally, 8625 images were obtained from the literature, and 517 SEM images describing the miscibility of blends were manually screened. Some images are from polymer blends with a small number of additives (generally below 3%), while the additives are basically not observed in the morphology. Researchers of these literature generally regard them as polymer blends and get miscibility judgement description according to the morphology. The miscibility labels of images were marked based on the PolyInfo and these descriptions in the literature. Since the descriptions of miscibility in literature and the PolyInfo database are mainly divided into miscibility and immiscibility, we treat miscibility prediction as a classification problem. The miscibility was labeled as miscible, immiscible, partially miscible. We also recorded the scale of SEM images of different polymers.

Table 1. **Statistics of Dataset**

| | Size | Split * | |
|---|---|---|---|
| | | Training-set | Test-set |
| Raw Image | 517 | 400 (78.8%) | 117 (86.3%) |
| Augmented | 2,082 | 1,965 (51.9%) | 117 (86.3%) |

* The figures in brackets mean the percentage of immiscible images.

In the dataset, the ratio of miscible images to immiscible images is about 1:4. Statistics of dataset are shown in Table 1. In the training set, there are 85 miscible images and 315 immiscible images. The imbalance of the dataset can result in bad performance of our model on miscible samples. To solve the problem of data imbalance and a small amount of data, we increased the amount of data in the training set through data augmentation to improve the generalization ability of the model. We augmented the miscible image by a factor of 12 and the immiscible image by a factor of 3 to try to balance both sides of the data as much as possible. Augmentation operations included rotation, translation, flipping, and finally cropping to a size of 256 * 256. Rotation means that we rotate the image at a random angle within 0 to 360 degrees. Translation means that we move horizontally and vertically image. Flipping means mirroring the image horizontally or vertically, where mirroring vertically is equivalent to rotating 180 degrees. Cropping means that we select a random 256*256 area from the image after rotation, translation and flipping. Finally, we got 1965 pieces of training data and 117 pieces of test data. We put specific examples in supporting information.

**2.2. Model**

Although many methods such as molecular dynamics, have been evaluated[29] and used in micro-structure analysis convolution neutral network[30] (CNN) is a better choice in automatic image recognition since the development of image datasets (such as ImageNet[27]) and computational power. Some kinds of model architectures with CNN module have been proposed. We designed a deep learning model based on VGG-Net[31] to meet our goals. At the same time, we applied some competing models to our dataset, including other deep learning algorithms such as Dense-Net[32] and Res-Net[33] and a traditional edge detection algorithm like Sobel. Details of these models are introduced as follows.

2.2.1. VGG-Net

CNN extracts different morphological features of the image through the convolution, and then pools the image, and finally gets the predicted result. VGG-Net was proposed by Oxford's Visual Geometry Group. VGG-Net uses two or three 2×2 or 3×3 convolution kernels to replace the larger convolution nuclei in the previous convolutional neural network. The stacked small convolution kernel can improve the depth of the network while maintaining the same receptive field, and more complex models can be learned through multi-layer nonlinear layers. Stacking small convolution kernels also reduces the number of model parameters and training costs. The VGG16 model (Figure 3A) used in this experiment contains 16 hidden layers (including 13 convolution layers and 3 fully connected layers), and the original 1000-dimensional soft-max is modified to 2-dimensional to describe the classification problem of miscibility.

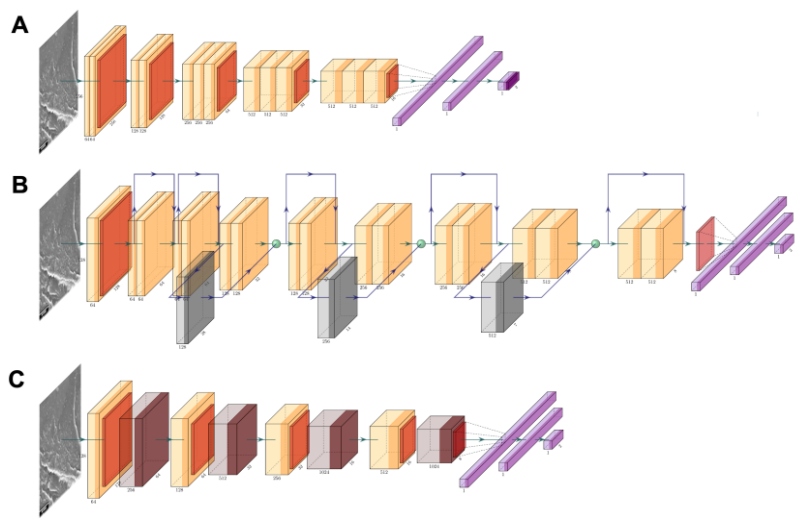

Figure 3. Diagram of CNN model used to predict polymer blends miscibility, in which yellow blocks represent the convolution layers, red blocks represent the pooling layers, and purple blocks represent the linear classification layer. In this experiment, the final classification layer was modified into 2-dimension to predicting the miscibility. A) VGG16; B) ResNet18, in which grey blocks represent shortcut connection block; C) DenseNet121, in which brown blocks represent Dense block.

2.2.2. Res-Net

It is widely believed that with the increase of the depth of the network, the fitting effect of the network on the function becomes better. However, Kaiming He et al. found that the accuracy of the model degrades with the deepening of the number of layers of the network, which they attribute to the fact that the depth of the network leads to the data being mapped to more discrete spaces, decreasing the fitting ability of simple mappings. So, they designed a ResNet model with "shortcut connection". Based on VGG model, ResNet added ResNet block, which changed the mapping from $F(x)$ to $F(x) + x$, improved the ability of the model to retain the identity transformation, so that the training of the deep model can be easier. The feature part of ResNet18(Figure 3B) is used to predict the polymer miscibility image, and the linear layer is modified.

### 2.2.3. Dense-Net

Based on ResNet, Gao Huang et al. proposed a densely connected mechanism and got the DenseNet model. Each layer in DenseNet connect with all previous layers in the channel dimension, so for a network of layer $L$, DenseNet contains $\frac{L(L+1)}{2}$ connections in total. Since dense connections require consistent feature map sizes, DenseNet uses the structure of DenseBlock + Transition. DenseBlock adopts a dense connection mode within layers and reduces the image size through the Transition layer. DenseNet can achieve better results than ResNet in some cases while the parameters are greatly reduced. The feature part of Dense-Net121(Figure 3C) is used to predict the polymer miscibility image, and the linear layer is modified.

### 2.2.4. Sobel

The Sobel edge detection algorithm is a method to detect image edges. The edge is the abrupt change in grayscale gradient, which usually corresponds to the phase interface between different polymers in SEM images. The Sobel operator is the core of Sobel edge detection, which includes two sets of 3×3 matrices, which monitor the horizontal edge and the longitudinal edge, respectively. The greatest difference between Sobel and machine learning methods above is that the Sobel operator is fixed while convolution cores in machine learning are learnable. By convolving the Sobel operator with the plane graph, the approximate values of the lateral and longitudinal brightness difference can be obtained, which shows the edges in the pictures[34] shows the raw image and the image processed by the Sobel algorithm.

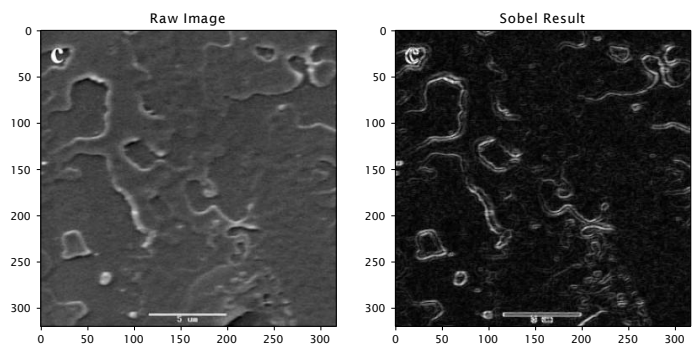

Figure 4. Comparison of Raw Image and Sobel Image.

Adapted with permission from reference 34. Copyright 2004 Society of Plastics Engineers

### 2.3. Transfer Learning

Transfer learning, as a widely used machine learning method, can significantly improve the performance of the model, especially in the fields of natural language processing and image recognition. Transfer learning aims to extract knowledge from source tasks and data (which is called 'pretrain'), and then apply that knowledge to target tasks (which is called 'finetune'). [35] Instead of learning the task from zero, transfer learning techniques try to transfer knowledge from the pretraining process when there is less high-quality training data for the target task. We applied transfer learning methods by utilizing the models pretrained with ImageNet images in advance and finetuning the classification parameters with the SEM images. It's assumed that the model can learn the patterns from millions of images in the pretraining process and converge faster on our data.

**2.4. Experiment Setting**

We conducted all experiments in Python using Pytorch toolbox, and the computing was accelerated on a NVIDIA GeForce RTX 2080 Ti GPU. We set the maximal training epoch number to 100 and the mini-batch to 128. The initial learning rate was set to $10^{-4}$. The total time spent was about 10 min for a complete training of 100 iterations. The Adam optimizer[36] was used to optimize the loss function. As for criterion, we chose cross-entropy as the loss function since it's a binary classification problem:

$$Loss = CrossEntropy(y, p) = \frac{1}{N}\sum_i -[y_i log(p_i) + (1 - y_i)log(1 - p_i)], \quad (1)$$

where $p_i$ represents the probability of the sample $i$ being miscibility-related.

Since we are accomplishing a classification prediction task, we also care about the accuracy of the model. In addition, we test some other indices:

$$Accuracy = (TP + TF)/(P + F), \quad (2)$$
$$Precision = TP/(TP + FP), \quad (3)$$
$$Recall = TP/(TP + FN), \quad (4)$$
$$Specificity = TN/(TN + FP), \quad (5)$$
$$F1\ score = 2 \times \frac{Precision \times Recall}{Precision + Recall}, \quad (6)$$

where $P$ means miscible, $N$ means immiscible, $T$ means true, and $F$ means false. Precision refers to the proportion of the "miscible" predictions that are truly miscible. Recall refers to the proportion of miscible samples with the correct prediction among all miscible samples. Specificity means the ability to predict immiscible cases. The $F1$ score comprehensively considers the effects of precision and recall. If one of them is too small, the value of $F1$ will become smaller.

**3. Results**

**3.1. Competing Model**

For each model, we conducted 20 experiments with different random seed. Besides, we invited 21 researchers to participate in our SEM images recognition. The researchers have adequate chemical knowledge and learned more details about polymer miscibility from literature and our dataset. The best results of the competing models and the average result of top-10 researchers are presented in Table 2. The accuracy of pretrained Dense-Net (91.45%) and pretrained Res-Net (90.60%) is close to that of VGG-Net (94.02%), which shows the effectiveness of the deep learning models and convolution networks. We find that, except for the Sobel method, all pretrained models performed well on the test set, which got even higher scores than the researchers' results. The reason may be that the details in SEM images are difficult for researchers to distinguish in some cases, so researchers have to make subject decisions. In addition, labels in our dataset are determined with SEM and other examinations (such as DSC) to guarantee the correctness while researchers can only utilize the microscopic information. Image recognition model can recognize all details in SEM images and learn the complex relation between SEM images and miscibility, so perform better than our researchers.

Moreover, the specificity of all models is much higher than their precision. As explained in experiment setting, precision means the ability to predict miscible cases while specificity means the ability to predict immiscible cases. Hence, for different categories, all models pay more attention to immiscible images. The reason may be the unbalanced proportion of training data. Although we augmented the training data, the raw images still influence the effect of training process. The number of immiscible images is nearly four times as large as the number of miscible images.

Table 2. Results of Competing Models

| Model | Test_Accu (%) | Train_Accu(%) | Precision (%) | Recall (%) | Specificity (%) | $F_1$ score (%) |
|---|---|---|---|---|---|---|
| Pre-VGG | **94.02** | 99.44 | **81.25** | **76.47** | 97.00 | **78.78** |
| Pre-Dense | 91.45 | 99.85 | 71.43 | 62.50 | 96.04 | 66.67 |
| Pre-Res | 90.60 | 99.19 | 69.23 | 56.25 | 96.04 | 62.07 |
| Sobel [1] | 80.20 | —— | 47.06 | 8.00 | **97.80** | 13.68 |
| Researcher | 83.88 | —— | 47.17 | 76.25 | 85.10 | 57.23 |

1. Boundary of classifier in Sobel method is set to 18 to achieve the highest accuracy.

### 3.2. Ablation Results

We have shown the effectiveness of several deep learning models with the pretrained method. To verify the importance of pretrain process, we conducted ablation experiments where pretrained parameters of models were not used. All models were trained on our SEM images dataset ab initio. The results are listed in Table 3. Besides, we showed the statistical data of all 20 experiments in Figure 5.

Table 3. Comparison Results of Pretrained (Pre) and No-Pretrained (Nopre) Models

| Model | Test_Accu (%) | Train_Accu (%) | Precision (%) | Recall (%) | Specificity (%) | $F_1$ score (%) |
|---|---|---|---|---|---|---|
| Nopre-VGG | 87.18 | 55.47 | 100.00 | 6.25 | 100.00 | 11.76 |
| Pre-VGG | **94.02** | 99.44 | 81.25 | 76.47 | 97.00 | **78.78** |
| Nopre-Dense | 88.89 | 68.04 | 63.63 | 43.75 | 96.04 | 51.85 |
| Pre-Dense | **91.45** | 99.85 | 71.43 | 62.50 | 96.04 | **66.67** |
| Nopre-Res | **91.45** | 97.30 | 71.43 | 62.50 | 96.04 | **66.67** |
| Pre-Res | 90.60 | 99.19 | 69.23 | 56.25 | 96.04 | 62.07 |

It is clear that VGG-Net and Dense-Net both performed better when pretraining was conducted; they improved by 7% and 3%, respectively. It is also indicated in Figure 5. When models were no-pretrained, it's difficult to train them with only our dataset, and their accuracy is not higher than ratio of immiscible images in both training set and test set. It means they did not learn the correct rule and just assumed all images immiscible. We took the whole results in the training process and compare the models with and without pretrained method, and presented them in Figure 6.

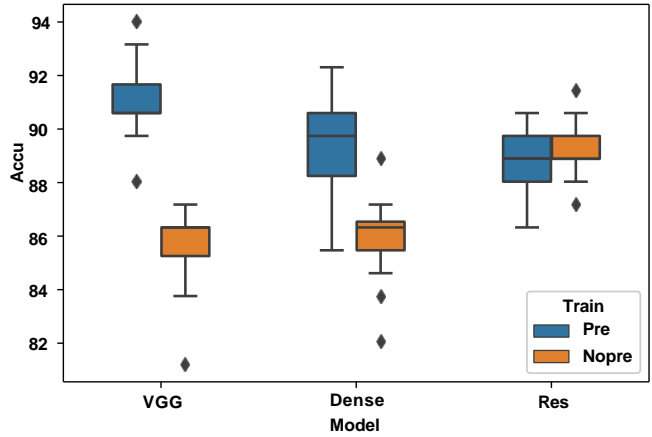

Figure 5. Box plot of the best accuracy in 20 experiments.

Average accuracy of pretrained method is higher than no-pretrained method for each model, which indicates the effectiveness of transfer learning.

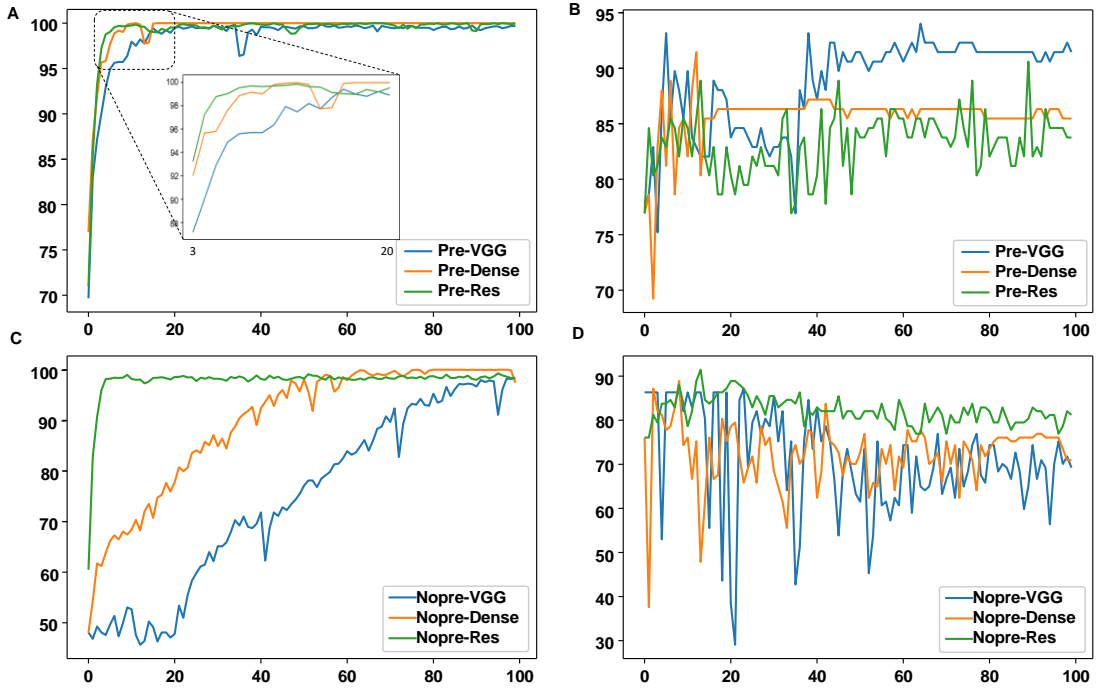

Figure 6. Accuracy of models on training set and test set in the training process. Abscissa means epochs.

(A) Accuracy of pretrained models on training set.

(B) Accuracy of pretrained models on test set.

(C) Accuracy of no-pretrained models on training set.

(D) Accuracy of no-pretrained models on test set.

As presented in Figure 5, we can find several interesting results. As to the training set, a pretrained model can rapidly converge and reach nearly 100% accuracy, while a no-pretrained model has to experience a relatively long process. This indicates that through pretraining, models reserved some basic knowledge so that when they were transferred to a new dataset, they could rapidly adjust parameters to solve the problem. As to the test set, we can also find that the pretrained models have a clearer and steadier optimizer process, while no-pretrained models fluctuate wildly. These results prove the effectiveness of the transfer learning method.

Furthermore, we can see that the rates of convergence on the training set differ during these untrained models. Res-Net without pretraining can converge rapidly on the training set, which is close to pretrained Res-Net, while no-pretrained Dense-Net and VGG-Net converge much more slowly than pretrained models. The difference may come from the model architecture and the connections between different layers in Res-Net and Dense-Net. The difference is connected with the result on the test set. We can find that no-pretrained Res-Net performs better than pretrained Res-Net. Meanwhile, the convergence rate gap between the no-pretrained and pretrained methods of Dense-Net and Res-Net is consistent with prediction accuracy on the test set. (2% for Dense-Net and 7% for VGG-Net)

**3.3. Case Study and Quantitative Criterion**

In order to investigate the methodology of our prediction model, we conducted a case study. We randomly chose 8 SEM images[37, 38] in our dataset and the best prediction model. We processed these images using the best model and obtained the prediction results. Labels and predictions for these images are listed in Table 4. Besides, we processed these images through Sobel method and got corresponding results (A1s-A4s and B1s-B4s).

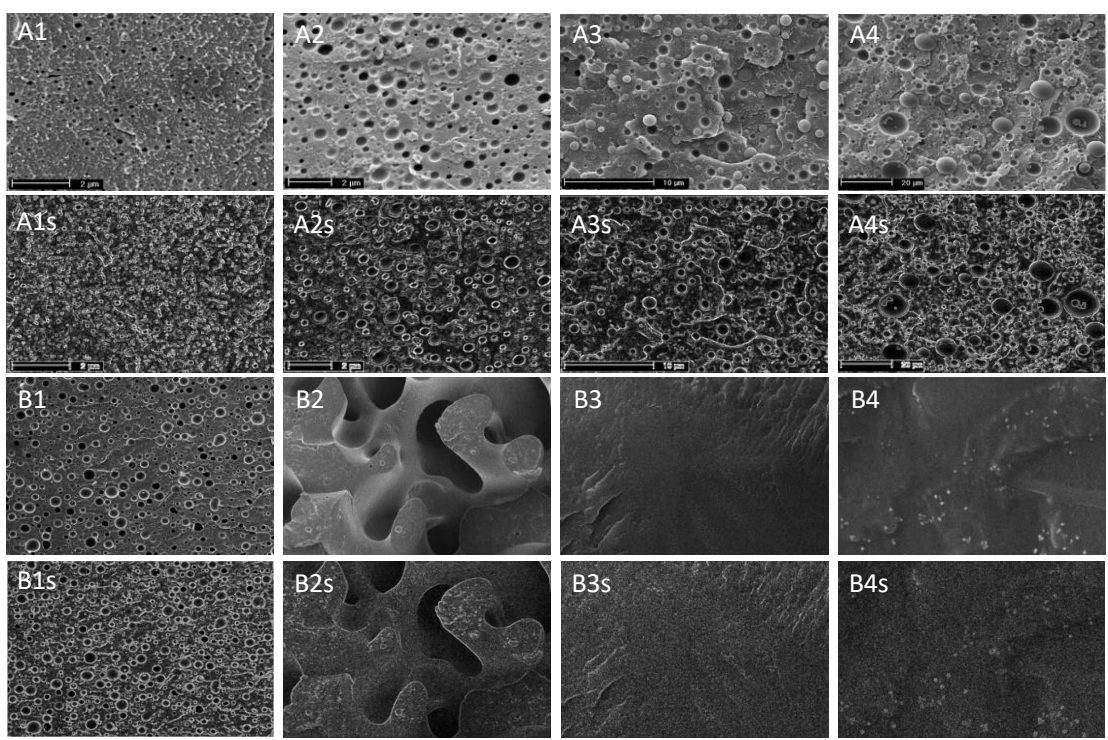

Figure 7. SEM images for case study.

A1-A4 show micrographs of the PMMA/PS1 and PMMA/PP blends. Adapted with permission from reference 37. Copyright 2005, Elsevier.

B1-B4 show SEM pictures of annealed and unannealed blends of PTT/PC. Adapted with permission from reference 38. Copyright 2009, American Chemical Society.

A1s-A2s and B1s-B4s show the results of Sobel method.

From Figure 7, we can find that Sobel method successfully detects the main edges in SEM images. However, only through Sobel method, lots of small particles are detected. After processed, the mean edge value of miscible samples can even much higher than immiscible samples. It means the limitation of the Sobel method when applied to miscibility recognition. From Table 4, we can find that the prediction result is consistent with the miscibility information. For SEM images A1-A4, the samples are all immiscible, although the degree of immiscibility is different. From these images, the particle diameter is listed in Table 4. From A1 to A4, the diameter increases, which indicates that the immiscibility increases. Through our model, we obtained the output for each image, including two categories. The final prediction is the SoftMax result of the two outputs, which means the probability of immiscibility. We can find that the prediction value also increases from A1 to A4. This result is consistent with the degree of immiscibility.

Table 4. Results of Case Study

| Image | Label | Particle Diameter | Output for Immiscible | Output for Miscible | Prediction for Immiscibility[1] |
|---|---|---|---|---|---|
| A1 | Immiscible | 0.24 μm | 0.08 | -30 | $e^{0.08}/(e^{0.08} + e^{-30})$ |
| A2 | Immiscible | 0.50 μm | 0.22 | -47 | $e^{0.22}/(e^{0.22} + e^{-47})$ |
| A3 | Immiscible | 1.49 μm | 0.48 | -47 | $e^{0.48}/(e^{0.48} + e^{-47})$ |
| A4 | Immiscible | 2.28 μm | -0.29 | -124 | $e^{-0.29}/(e^{-0.29} + e^{-124})$ |
| B1 | Immiscible | — | 0.23 | -67 | $e^{0.23}/(e^{0.23} + e^{-67})$ |
| B2 | Immiscible | — | 0.16 | -89 | $e^{0.16}/(e^{0.16} + e^{-89})$ |
| B3 | Miscible | — | -41 | 10 | $e^{-41}/(e^{-41} + e^{-10})$ |
| B4 | Miscible | — | -32 | 8 | $e^{-32}/(e^{-32} + e^{8})$ |

1. Prediction for Immiscibility is SoftMax result of Output for Immiscible and Miscible

As for B1-B4, these images include both miscible and immiscible samples. The particle information is not clear, but we can compare the outputs and predictions for different categories. The outputs for immiscible are both negative for B1 and B2, while positive for B3 and B4. It is obvious the predictions for immiscibility are all consistent with the labels. Through the case study above, the effectiveness of our model is shown, the quantitative criterion for polymer miscibility is presented.

## 4. Conclusion

In this work, we presented a machine learning framework to bridge material microscopic images and macroscopic properties. Start with polymer miscibility recognition, machine learning method was introduced into the materials field. We achieved 94% accuracy and presented a quantitative criterion for miscibility and conduct a case study to show its effectiveness.

Our research also yielded some intriguing results. Different deep learning models perform differently in the pretraining process, and it shows consistency with the final accuracy on the test set. As to the quantitative criterion, the result now shows polarization. Quantitative result of miscible sample is very close to '1' while it is close to '0' for immiscible samples.

Our method may still have some limitations, which can be improved in the future work. The performance of this method is constrained by the quality and clarity of SEM images. In SEM images, the convex of fracture surface is similar to the phase separation, which makes it different to distinguish its miscibility. Besides, more complicated cases (e.g. ternary or quaternary blends) exist in many practical applications. Since we collected polymer blend SEM image from literature and PolyInfo, our database mainly includes binary polymer blend cases. Only 18 SEM images of ternary blends exist in our database and the pretrained VGG achieved 94.4% classification accuracy on them. In the future, with the increase of dataset, we expect that our model can perform better in non-ideal situations.

    This work is an attempt to make the analysis of characterization results more automatic and intelligent. For example, it is possible for researchers to explore predict mechanical properties through SEM images in addition with other characterization results. In addition, other chemical examination results can be analyzed through machine learning method, such as infrared spectroscopy etc. We believe that it will inspire more creative methods and promote the development of the interdiscipline.

**Data and Software Availability**

The data and main code to reproduce the results of this study are available at following GitHub page: https://github.com/Zhilong-Liang/Polymer-Miscibility-Image-Recognition.

**Supporting Information**

Supplemental figure describing the confusion matrix of several models on the test set and figure of violin plot of result of Sobel method.

Author Contributions

Z.L., C.Z. and J.Y. conceived of the main research idea and carried out the method design and ML modeling. Z.T. took part in the literature summary and database building. R.H. and W.O. took part in the database building and revised the manuscript. C.Z. and J.Y. supervised the project and revised the manuscript. All authors discussed the results and commented on the manuscript.

Notes

The authors declare no competing financial interest.

Acknowledgement

This work is supported by a grant (no. 2021GQG1025) from the Guoqiang Institute, Tsinghua University, Beijing, China.

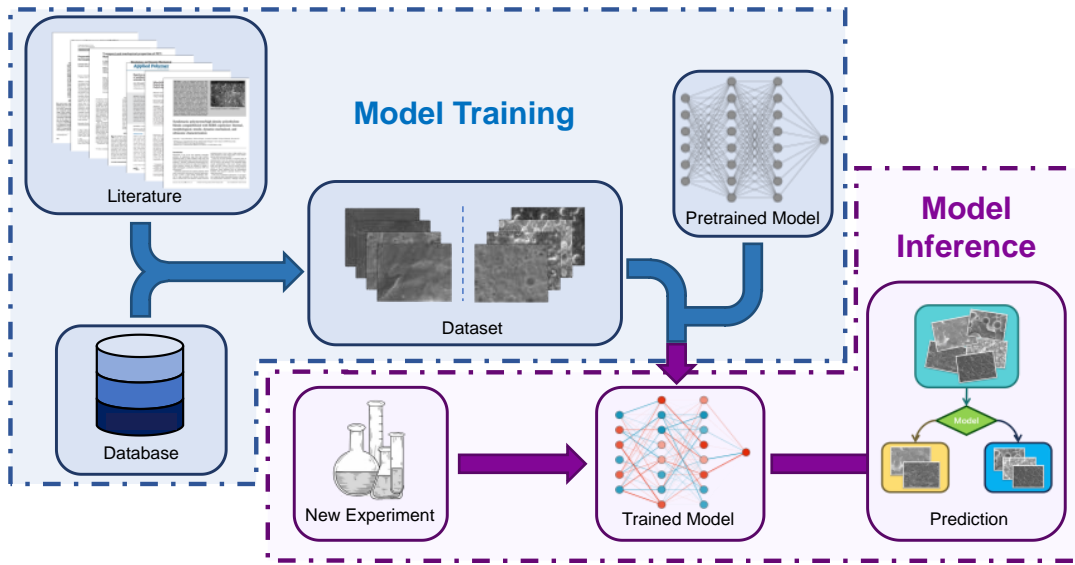